\documentclass[preprint,12pt]{elsarticle}

\usepackage{amssymb}
\usepackage{amsmath} 

\usepackage{tikz}
\usetikzlibrary{positioning}
\usetikzlibrary {arrows.meta}
\usepackage{makecell} 
\usepackage{colortbl} 
\usepackage{bbm} 
\usepackage{url}
\usepackage{xspace}
\usepackage{booktabs}

\journal{Pattern Recognition}

\begin{document}

\begin{frontmatter}



\title{Boosting CNN-based Handwriting Recognition Systems with Learnable Relaxation Labelling}

\author[1,2]{Sara {Ferro}\fnref{fnlab}}\ead{Sara.Ferro@iit.it}
\author[3,4]{Alessandro {Torcinovich}\fnref{fnlab}}\ead{alessandro.torcinovich@inf.ethz.ch}
\author[2,1]{Arianna {Traviglia}}\ead{Arianna.Traviglia@iit.it}
\author[1,2]{Marcello {Pelillo}\corref{corrlab}}\ead{pelillo@unive.it}

\cortext[corrlab]{Corresponding author.}
\fntext[fnlab]{Equal contribution, authors listed in alphabetical order}

\affiliation[1]{organization={DAIS, Ca' Foscari University of Venice},
            addressline={via Torino 155}, 
            city={Venice},
            postcode={30172}, 
            country={Italy}}
\affiliation[2]{organization={CCHT, Italian Institute of Technology},
            addressline={via Torino 155}, 
            city={Venice},
            postcode={30172}, 
            country={Italy}}
\affiliation[3]{organization={D. Eng., Free University of Bozen-Bolzano},
            addressline={Piazza Domenicani 3}, 
            city={Bolzano},
            postcode={39100}, 
            country={Italy}}
\affiliation[4]{organization={DINFK, ETH Zurich},
            addressline={Andreasstrasse 5}, 
            city={Zurich},
            postcode={8050}, 
            country={Switzerland}}

\begin{abstract}
The primary challenge for handwriting recognition systems lies in managing long-range contextual dependencies, an issue that traditional models often struggle with. To mitigate it, attention mechanisms have recently been employed to enhance context-aware labelling, thereby achieving state-of-the-art performance. In the field of pattern recognition and image analysis, however, the use of contextual information in labelling problems has a long history and goes back at least to the early 1970's. Among the various approaches developed in those years, Relaxation Labelling (RL) processes have played a prominent role and have been the method of choice in the field for more than a decade. Contrary to recent transformer-based architectures, RL processes offer a principled approach to the use of contextual constraints, having a solid theoretic foundation grounded on variational inequality and game theory, as well as effective algorithms with convergence guarantees.
In this paper, we propose a novel approach to handwriting recognition that integrates the strengths of two distinct methodologies. In particular, we propose integrating (trainable) RL processes with various well-established neural architectures and we introduce a sparsification technique that accelerates the convergence of the algorithm and enhances the overall system's performance. Experiments over several benchmark datasets show that RL processes can improve the generalisation ability, even surpassing in some cases transformer-based architectures.
\end{abstract}


\end{frontmatter}


\section{Introduction}
\label{sec:introduction}

Handwritten text recognition (HTR) is a core area in machine learning and pattern recognition, focused on identifying and interpreting handwritten characters within image data. HTR has countless applications in various fields, including document digitalisation and archiving, automated form processing, educational technology, information forensics, and more. Our interest in this problem stems from the  study of historical handwriting, specifically the need to convert ancient handwritten text from historical codices into a machine-readable format. This digitalisation effort aims to create publicly accessible archives with editable content, offering palaeographers and humanities scholars an invaluable resource.

Despite significant advancements in this field, HTR continues to pose several challenges, primarily due to the vast variability in character shapes, which are influenced by the writer's handwriting style, and the type of writing tools used. In addition, transcribing historical manuscripts is particularly difficult as the writing medium (\textit{e.g.}, paper or parchment) often deteriorates over time, further complicating the process.

Over the years, several approaches have been proposed to address the HTR problem.
Here, we briefly mention the most important ones and we refer the reader to \citep{2020LombardiFDeep,2022NTeslyaDeep} for a comprehensive review.
Traditionally, the input data has been processed by recurrent network models, taking into account one-directional \citep{1993SeniorAWAn}, bi-directional \citep{2005GravesABidirectional}, or multi-directional \citep{2008GravesAOffline} information. Subsequent findings have demonstrated that hybrid networks, specifically CRNNs, which integrate both convolutional and recurrent layers, achieve superior performance with fewer parameters \citep{2017PuigcerverJAre,2019CoquenetDHave,2022RetsinasGBest}. Subsequently, Fully Convolutional Networks (FCN) have been proposed to decrease the models' parameter count \citep{2020CoquenetDRecurrence} and also Gated Convolutional Neural Networks (GCNN) \citep{2020YousefMAccurate}, which try to filter the information flow to enable only pertinent information to pass. All the aforementioned models rely on the Connectionist Temporal Classifier (CTC) loss \citep{2006GravesAConnectionist}, which handles the sequence alignment issue introduced by the variability of the input text image and the input text itself. 

The main conceptual challenge for HTR systems stems from the difficulty of dealing with long-range contextual dependencies, a problem classical recurrent models often struggle with. To effectively deal with this problem and to deal more effectively with contextual information, recent state-of-the-art architectures utilise attention mechanisms \citep{2017BlucheTScan, 2022KangLPay, 2021LiMTrocr}. Here, the CTC loss is substituted by a cross-entropy term, as the alignment is tackled by an attention-based encoder-decoder architecture. 

The importance of contextual information in pattern recognition, however, has been recognised since the beginnings of the field and over the years several solutions have been proposed (\textit{cf.} \cite{Tou78} for a classical review). Starting from heuristic, {\em ad hoc} solutions, often motivated precisely by text recognition problems, the community gradually tried to develop more formal frameworks which could ideally encompass different kinds of contextual classification problems (\textit{e.g.}, \cite{HarSha79}). 
These efforts resulted eventually in the development of \emph{Relaxation Labelling (RL)} processes \cite{1976RosenfeldAScene} and, later, in a now classical theory of consistency \cite{1983HummelRAOn}. Since their introduction, RL and similar processes have played a prominent role in the fields of pattern recognition and image analysis and have been the method of choice for more than a decade \citep{2001ZuckerSRelaxation}.

These algorithms work as dynamical systems, employing contextual information to enhance the accuracy of labelling assignments. Similar to attention-based models, which employ self-attention \citep{2017VaswaniAAttention}, RL enables the message-passing of information among fundamental elements within a given context (\textit{e.g.}, characters in a text), determining the most suitable labelling that aligns with the data configuration. In particular, RL considers \emph{compatibilities} between (or among, in case of high-order constraints) labelling hypotheses in an attempt to refine an initial labelling assignment until it reaches a final \emph{consistent labelling} which adheres to the (soft) constraints expressed by the compatibility function. Differently from self-attention, RL offers well-established theoretical convergence properties \cite{1997PelilloMThe} and solid mathematical foundations grounded on variational inequality theory and ultimately game theory \cite{1983HummelRAOn,1991MillerDACopositivePlus}. 

Not surprisingly, RL processes have already been used in the context of text recognition (\textit{e.g.,} \citep{1988GoshtasbyAContextual}), using handcrafted compatibility functions. However, these compatibilities demand domain knowledge of the specific problem and usually lead to poor adaptability when the nature of the data source changes. On the contrary, several works have clearly demonstrated the possibility of efficiently learning similarity metrics associated with meaningful embedding spaces \citep{2013QiongCSimilarity, 2020EleziIThe}. In the case of RL processes, a classical forward-propagation strategy has been shown to be able to effectively learn compatibility functions from data \citep{1994PelilloMLearning}.

In a previous preliminary work \citep{2023FerroSExploiting}, we incorporated learnable RL processes into a specific realisation of a Convolutional Recurrent Neural Network (CRNN) and we managed to improve its overall performance. In particular, we replaced the forward-propagation strategy alluded to above with the standard backward propagation scheme in an attempt to achieve end-to-end learning of the parameters of both RL and the neural backbone. Motivated by the promising results of our previous work, in this paper, we broaden our study by applying RL to various well-established NN-based HTR systems. Furthermore, we introduce a sparsification procedure for the compatibility coefficients which allows us to speed up the convergence of the processes and enhances the system's overall performance. We conducted experiments across several HTR datasets, demonstrating that RL processes can improve the generalisation capability across different baselines, even surpassing the performance of much larger state-of-the-art transformed-based architectures.

\section{Relaxation Labelling Processes}
Originated in the context of image analysis and computer vision, relaxation labelling processes aim to solve \emph{consistent labelling problems}, namely problems where one has to assign labels to objects in a way that adheres (or is ``consistent'' with) problem-specific contextual constraints \cite{1976RosenfeldAScene,HarSha79}.
These constraints can be given or learned from data, as in \cite{1994PelilloMLearning}.

Attempts at formalising the notion of a consistent labelling culminated in a seminal work by Hummel and Zucker \cite{1983HummelRAOn}, who developed a formal theory of consistency based on variational inequality theory that later turned out to have intimate connections with non-cooperative game theory \cite{1991MillerDACopositivePlus}. The theory generalises the classical constraint satisfaction problem (which uses Boolean constraints) to ``soft'' compatibility measures and probabilistic labelling assignments \cite{1976RosenfeldAScene}. 

More formally, suppose that a set of objects $B = \left\{b_1, \dots, b_n\right\}$ and a set of labels $\Lambda = \left\{1, \dots, m\right\}$ are given. The aim is to label each object of $B$ with a label in $\Lambda$, and we try to accomplish this by exploiting two sources of information. 
One is \emph{local} information, and captures the salient features of each individual object taken in isolation (this is then encoded in the prior, or initial, distribution, as described later). The other is \emph{contextual} information, which  
takes into account the agreement among different object-label hypotheses. This agreement is quantitatively expressed in terms of \emph{compatibility coefficients}. 

Typically, these coefficients express the compatibility between pairs of hypotheses (but see \cite{1983HummelRAOn,1997PelilloMThe} for high-order generalisations), and hence can be organised in terms of a matrix $R$ composed of $n \times n$ blocks:
\begin{equation}\label{eq:R_def}
    R = \begin{bmatrix}
R_{11} & \hdots & R_{1n}\\
\vdots & \ddots & \vdots\\
R_{n1} & \hdots & R_{nn}
\end{bmatrix},
\end{equation}
where each $R_{ij}$ is a $m \times m$ matrix:
\begin{equation}\label{eq:R_submat_def}
        R_{ij}=\begin{bmatrix}
r_{ij}(1,1) & \dots & r_{ij}(1,m)\\
\vdots & \ddots & \vdots \\
r_{ij}(m,1) & \dots & r_{ij}(m,m)
\end{bmatrix}.
\end{equation}
Each coefficient $r_{ij}(\lambda, \mu) \ge 0$ measures the strength of compatibility between the hypotheses ``$\lambda$ assigned to $b_i$'' and ``$\mu$ assigned to $b_j$.'' High values correspond to agreement in the hypotheses, low values to disagreement.

Let $p_i(\lambda)$ denote the probability that object $b_i$ is labelled with label $\lambda$. 
An RL algorithm starts with an $m$-dimensional prior for each object $i \in B$:
\begin{equation}
    p_i^{(0)}= (p_i^{(0)}(1),\dots, p_i^{(0)}(m)),
\end{equation}
with $p_i^{(0)}(\lambda)\geq 0$ and $\sum_\lambda p_i^{(0)}(\lambda)=1$, for $i=1,\dots , n$, 
and iteratively refines it taking into account contextual constraints. This way, both local and contextual information contribute to the final object-label assignment and there is a clear division of labour: local information (obtained, for example, by standard feature extraction methods) provides the starting point of the algorithm, while contextual constraints are used in the refinement process.

Note that this differs markedly from algorithms based on Markov Random Fields (MRF's) and related approaches \cite{Li2009}, whereby the labelling problem is cast in terms of finding a (global) minimiser of an objective function consisting typically of two terms: a unary term, which encodes prior/local information, and a quadratic (or possibly higher-order) term encoding contextual constraints. On the contrary, in RL processes there is no attempt at solving a global optimisation problem as the prior information defines the starting point of a dynamical system and the final goal is to converge to the ``closest'' consistent labelling assignment. In fact, convergence to a global solution would mean that the algorithm has ``forgotten'' the prior information, which is of course an undesirable property (for more details on this point, see the discussion contained in \cite{1983HummelRAOn}). As a consequence, one of the most appealing aspects of RL processes is that they avoid the common and difficult problem of finding global optima, which is a challenge in most optimisation tasks.

In RL processes each object is associated with an initial probability distribution, and the concatenation of all these distributions forms a \emph{weighted labelling assignment}, that is an $nm$-dimensional vector\footnote{Of course, a weighted labelling assignment can also be thought of as a stochastic matrix, but it is mathematically more convenient to consider it as a vector.} ${\bf p}^{(0)}$. The set of all possible weighted labelling assignments is denoted by $\mathbb{K}$:
\begin{align}
    \begin{aligned}[t]
        \mathbb{K} = \{ {\bf p} \in \mathbb{R}^{nm} \mid &\  p_i(\lambda) \geq 0 \mbox{  and  } \textstyle\sum\nolimits_{\lambda} p_i(\lambda)=1,\\
        & \ i = 1, \dots, n,\ \lambda \in \Lambda \}.
    \end{aligned}
\end{align}

Now, given a weighted labelling assignment $\bf p \in \mathbb{K}$, the quantity
\begin{equation}
    q_i(\lambda)=\sum_j\sum_\mu r_{ij}(\lambda, \mu)p_j(\mu)
\label{eq:relaxation_op}
\end{equation}
measures the \emph{support} that context gives to the hypothesis ``object $b_i$ is labelled with label $\lambda$''.
Motivated by the theory of variational inequalities, Hummel and Zucker \cite{1983HummelRAOn} defined $\bf p$ to be \emph{consistent} if, for all $i = 1, \dots, n$:
\begin{equation}
    \sum_{\lambda = 1}^m p_i(\lambda)q_i(\lambda) \ge \sum_{\lambda}^m p_i^\prime(\lambda)q_i(\lambda)
\end{equation}
for all ${\bf p}^\prime \in \mathbb{K}$. Geometrically, this means that the support vector $\bf q$, obtained by putting together all the $q_i(\lambda)$'s, points away from all tangent directions.  

The classical RL algorithm introduced in \cite{1976RosenfeldAScene}, which is the one used in this paper, takes as input the initial labelling assignment ${\bf p}^{(0)}$ and produces a sequence of labellings ${\bf p}^{(1)}, {\bf p}^{(2)}, \ldots \in \mathbb{K}$
using the following update rule: 
\begin{equation}
    p_i^{(\tau+1)}(\lambda)=\frac{p_i^{(\tau)}(\lambda)q_i^{(\tau)}(\lambda)}{\sum_{\mu}p_i^{(\tau)}(\mu)q_i^{(\tau)}(\mu)}
\label{eq:next_p}
\end{equation}
with $\tau = 0, 1, 2, \ldots$ indicating the iteration step. 
In theory, the process should proceed until it reaches a fixed point, namely until ${\bf p}^{(\tau+1)}={\bf p}^{(\tau)}$ for some $\tau$.
In practical applications, however, it is typically stopped either when the distance between two consecutive labellings becomes negligible, or after reaching a predetermined number of steps.

Although originally developed in a purely heuristic manner, Pelillo \cite{1997PelilloMThe} showed that this dynamical system turns out to have an intimate connection with Hummel and Zucker's consistency theory.
In fact, under the assumption of symmetry of the compatibility matrix $R$, the process is proven to converge to (local) maximizers of the so-called \emph{average local consistency}
\begin{equation}
    A(p) = \sum_{i}\sum_{\lambda}p_i(\lambda)q_i(\lambda),
\end{equation}
which in this case are known to correspond to consistent labellings \cite{1983HummelRAOn}.
Similar (but weaker) convergence properties also hold in the case of asymmetric compatibilities \cite{1997PelilloMThe}.

\section{Integrating Relaxation Labelling with CNN's}\label{sec:prop_model_and_learn_scheme}
In this section, we describe how to integrate trainable RL processes with various neural network architectures in order to improve the accuracy of HTR systems. Although consistent labellings are not guaranteed to maximise standard HTR performance metrics, in the experimental section we show empirically that this is indeed the case, thereby confirming the benefits of pursuing a formal agreement among labelling hypotheses. In so doing, the RL processes provide a principled way to capture informative long-range relationships among textual tokens in their respective context. We shall consider both recurrent (CRNN) and fully convolutional (FCN) architectures.

The proposed combined models consist of a neural-network backbone (referred to as the \emph{baseline} in the sequel) and an RL module which refines the baseline predictions before the recurring module (if present), to avoid running into missing contextual information that can occur when using recurring layers \cite{1994BengioYLearning}. 
In the case of the FCN, the RL is placed before the decoder module to maintain a similar architecture to the case of having the recurrent module (see below for details).

\subsection{The Baseline Models}
Three different state-of-the-art CRNN models and one state-of-the-art FCN architecture are considered as the baselines. 
As for the CRNN's, we considered the models developed by
Shi \textit{et al.} \cite{2016ShiBAn},
Puigcerver \cite{2017PuigcerverJAre}, and
Retsinas \textit{et al.} \cite{2022RetsinasGBest}, 
the main differences being the layer depths and the composition of the convolutional module. The FCN architecture considered is the one developed by Coquenet \textit{et al.} \cite{2022CoquenetDEnd}.

Shi \textit{et al.} \cite{2016ShiBAn} proposed a convolutional module based on VGG-$11$ \cite{2014SimonyanKVery}, with the addition of BatchNorm \cite{2015IoffeSBatch}, and an ending convolutional block. Furthermore, the pooling layers are changed to have stride to $1 \times 2$, to accommodate for the text data. The recurrent part consists of $2$ BLSTM layers. The model introduced in Puigcerver \cite{2017PuigcerverJAre} contains $5$ convolutional blocks, comprising BatchNorm, LeakyReLU and Max-pooling. The recurrent part is composed of $5$ BLSTM layers. The last CRNN architecture considered is the one of Retsinas \textit{et al.} \cite{2022RetsinasGBest} which has a deep convolutional part featuring one convolutional block followed by $10$ ResNet blocks \cite{2015HeKDeep} with ReLU, BatchNorm, dropout and Max-pooling. The recurrent part comprises $3$ BLSTM layers. Differently from the other models, the connection between the convolutional and the recurrent modules is established through column-wise max-pooling instead of column-wise concatenation. In addition, a \emph{CTC shortcut} is used, consisting of a convolutional layer with kernel size $1 \times 3$, used to connect the convolutional part to another CTC term.

The FCN model of Coquenet \textit{et al.} \cite{2022CoquenetDEnd} used in this study is composed of an encoder of $6$ convolutional blocks with $16$, $32$, $64$, $128$ and $128$ channels. These are followed by $4$ different Depth-Wise Separable Convolutions (DWSC), where the first $3$ have $128$ channels and the last $256$ channels. Between the DWSC, there are skipping connections. Finally, the decoder is connected to the encoder through an Adaptive Max-pooling. It is composed of a convolution presenting the number of the output channels equal to the number of characters in the alphabet (comprising the blank character needed for the CTC loss \cite{2006GravesAConnectionist}).

For all the models, a fully connected layer is used for mapping the output dimension of the decoder to the number of characters in the alphabet (comprising the blank character). The scores are then reported to a probability distribution by using the SoftMax function \cite{1989BridleJTraining}.

\subsection{The Proposed Combined Architectures}

\begin{figure}[!t]
\centering
\includegraphics[width=\columnwidth]{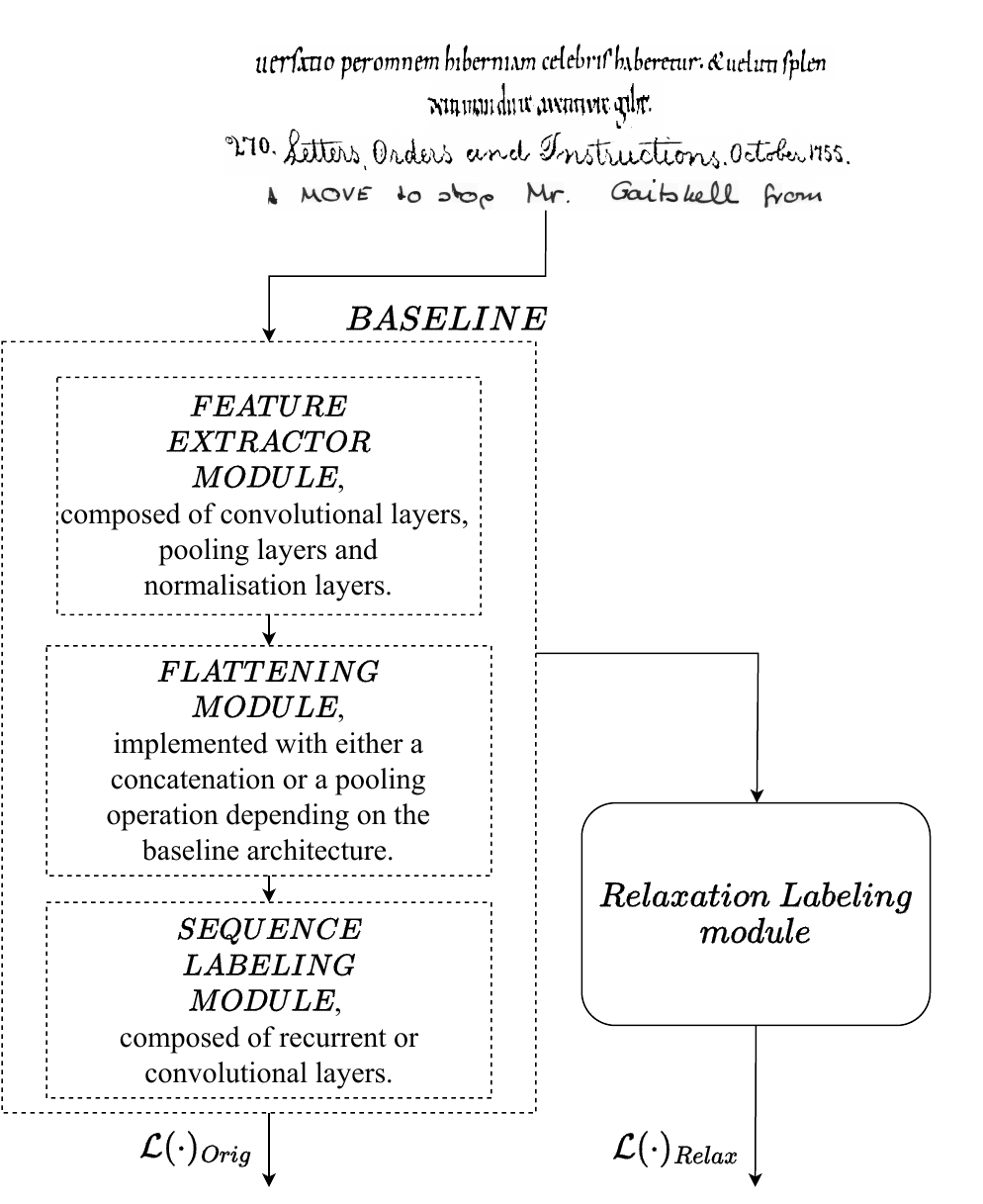}
\caption{Combined architecture of a baseline with the RL module (see text for details).}
\centering
\label{fig:comb_arch}
\end{figure}

Fig.~\ref{fig:comb_arch} shows the general structure of the proposed combined architectures. The RL module is applied in between the encoder and the decoder modules of the baseline. This decision is due to the fact that in CRNN models, the decoder can be affected by issues such as vanishing gradient. In particular, in the cases of Retsinas \textit{et al.} \cite{2022RetsinasGBest} and Coquenet \textit{et al.} \cite{2022CoquenetDEnd} the RL module is directly connected after the pooling layer that bridges the encoder and decoder. Differently, in Shi \textit{et al.} \cite{2016ShiBAn} and Puigcerver \cite{2017PuigcerverJAre}, the encoder is connected to the decoder through a flattening module performed by concatenating feature vectors. In such cases, we introduced a novel branch to connect the RL module before the concatenation. It consists of a max-pooling layer followed by a fully connected layer and a SoftMax activation, used to obtain the probability distribution in input to the RL. This approach is designed to minimise the number of parameters introduced by such a branch.

Considering the size of the compatibility matrix, the RL module is used only at training time as a regulariser, and it is then removed during inference time, using the baseline architectures only.

\subsection{End-to-end Learning}
To learn the parameters or the RL process, the backward propagation through time algorithm (BPTT) \cite{2013GuoJBackpropagation,1990WerbosPBackpropagation} is used, which is guaranteed to produce equivalent results with respect to the forward propagation learning scheme originally proposed in \cite{1994PelilloMLearning} but with the additional advantage of being computationally efficient \cite{1994BengioYLearning,2018BaydinAGAutomatic}.

\begin{figure*}[!t]
    \centering
    \includegraphics[width=0.70\textwidth]{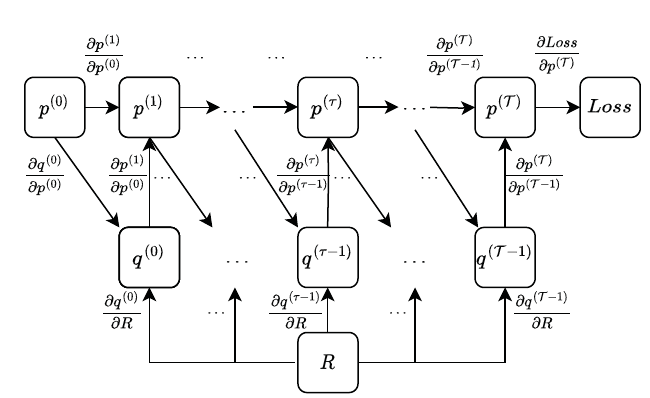}
    \caption{Computational graph depicting the process of error backward propagation in the RL module, highlighting its temporal or iterative nature through the RL process.}
    \centering
    \label{fig:bptt_relab}
\end{figure*}
Fig.~\ref{fig:bptt_relab} depicts the computational graph of the RL module for a fixed number of iterations $\mathcal{T}$. The iterations of the RL module can be conceptualised as a sequential advancement through time steps, and this progression can be visually represented with the computational graph.
In the same figure, we also present the derivatives relevant to BPTT.

The loss employed for all models under study is the Connectionist Temporal Classification (CTC) loss \cite{2006GravesAConnectionist}, which enables training with unsegmented data, solving the necessity to pinpoint precisely the word or character positions within the image.

The CTC loss is calculated between a continuous, unsegmented time series and a target sequence. In the context of this work, the time series is represented by the image of a handwritten text line, while the target sequence corresponds to its transcription. By summing the probabilities of all possible alignments between the input image and the target sequence, the probability of the target given the input can be computed. This results in a differentiable loss value with respect to each input neuron.

To learn the parameters of the combined architectures, the original loss of each model $\mathcal{L}_{Orig}$ is sided with a $\mathcal{L}_{Relax}$ term, computed as a CTC loss over the RL refined predictions. In addition, differently from previous work \cite{2023FerroSExploiting}, an $\ell_1$ regularisation term is added to sparsify the matrix of the compatibility coefficients and to contrast overfitting \cite{2017ChengYA}.

The total loss used to train the combined architecture is
\begin{equation}
    \label{eq:loss}
    \begin{split}
        \mathcal{L}\left(\rho_{conv}, \rho_{rec}, R; s\right) & =  \mathcal{L}_{Orig}(\rho_{conv}, \rho_{rec};s) \\
        & + \beta \mathcal{L}_{Relax}(\rho_{conv}, R;s)+\gamma \|R\|_1,
    \end{split}
\end{equation}
with $\rho_{conv}$, $\rho_{rec}$ representing respectively the parameters of the convolutional and recurrent parts\footnote{Note that the FCN architecture does not present a recurrent component. In this context, we present the general formula that encompasses all possible parameters.} of the baseline architecture, $R$ being the compatibility matrix, $s$ the transcription of the input line, $\beta$ and $\gamma$ are weighting hyperparameters.

As the $\mathcal{L}_{Relax}$ is a CTC loss, the derivatives with respect to the probabilities are the standard ones.
The attention is directly focused on the derivatives depending on the compatibility coefficients. Differently from \cite{1994PelilloMLearning}, equivariance and parameter sharing is not assumed.

Eq. \eqref{eq:next_p} can be written as $p^{(\tau+1)}_{i\lambda}=\frac{h_{i\lambda}^{(\tau)}}{\sum_\mu h_{i\mu}^{(\tau)}}$, where $h_{i\lambda}^{(\tau)} = p_{i\lambda}^{(\tau)}q_{i\lambda}^{(\tau)}$. From here, its derivative is
\begin{equation}
    \frac{\partial p_{i\lambda}^{(\tau + 1)}}{\partial h_{i\eta}^{(\tau)}} = \left(\mathbbm{1}(\lambda = \eta)\sum_{\mu}h_{i\mu}^{(\tau)} - h_{i\lambda}^{(\tau)}\right)\text{\LARGE$/$}\left(\sum_{\mu}h_{i\mu}^{(\tau)}\right)^2,
\end{equation}
where $\mathbbm{1}(\cdot)$ is the indicator function. The derivatives of $h(\cdot)$ are respectively $\frac{\partial{h_{i\lambda}^{(\tau)}}}{\partial {p_{i\lambda}^{(\tau)}}}= q_{i\lambda}^{(\tau)}$ and $\frac{\partial h_{i\lambda}^{(\tau)}}{\partial {q_{i\lambda}^{(\tau)}}}=p_{i\lambda}^{(\tau)}$.\\
Finally, the derivative of $q_{i\lambda}^{(\tau)}$ over $r_{hk\alpha\beta}$ is
\begin{align}
        \frac{\partial q_{i\lambda}^{(\tau)}}{\partial r_{hk\alpha\beta}}
        & = \sum_{j=1}^n\sum_{\mu=1}^m \mathbbm{1} \left(h = i, \alpha = \lambda \right)p_{j\mu}^{(\tau)} + r_{hk\alpha\beta} \frac{\partial p_{j\mu}^{(\tau)}}{\partial r_{hk\alpha\beta}}.
\end{align}

\section{Experimental Setting}\label{sec:exp}
This section provides an overview of the datasets used and other details related to the experiments.

\subsection{Datasets}
We used both historical and modern datasets with line-level transcription\footnote{Datasets available at: \url{https://fki.tic.heia-fr.ch/databases}.}: the Saint Gall \cite{2011FischerATranscription}, the Parzival \cite{2012FischerALexicon-free}, the Washington \cite{2012FischerALexicon-free}, and the IAM dataset \cite{2002MartiUVThe}\footnote{Adopted splitting: \url{http://www.tbluche.com/resources.html}. The dataset partition consists of $6482$ training samples, $976$ validation samples, and $2915$ test samples.}. Saint Gall comprises manuscripts from the $9^{th}$ century written in Latin, while Parzival has manuscripts from the $13^{th}$ century written in German. Washington contains handwritten letters in English from the $18^{th}$ century. Finally, IAM comprehends forms of handwritten modern English text from $657$ different writers. We adopted the partition in \cite{2014PhamVDropout}, and this study does not compare with methodologies that use different splittings (\textit{e.g.},  Michael \textit{et al.} (2019) \citep{2019MichaelJEvaluating}, Yousef \textit{et al.} (2020) \citep{2020YousefMAccurate}, and Diaz \textit{et al.} (2021) \citep{2021DiazDHRethinking}). Transcription errors are present in all the datasets, a well-known problem that is affecting the performance of the models \cite{2020AradillasJImproving}.

\subsection{Data Pre-processing and Augmentation}
We applied the pre-processing practices proposed in \cite{2022RetsinasGBest}. In particular, we performed image centering and left and right padding with median intensity. Furthermore, we used classical data augmentation techniques such as random affine transformations as rotations and translations to increase the number of samples. Only in the case of the IAM dataset, we utilised a Gaussian blur filter of kernel $3 \times 3$ with a randomly chosen standard deviation $\sigma \in \left[1, 2\right]$. The images were resized to $128 \times 1024$ ($H \times W$). All data pre-processing and augmentations were kept the same across the architectures, for a fair comparison.

\subsection{Settings}
\begin{table*}[!t]
\centering
\begin{tabular}{lccc}
  \hline
  \cellcolor{gray!25} \small{Model authors} & \cellcolor{gray!25} \small{Saint Gall} & \cellcolor{gray!25} \small{Parzival} & \cellcolor{gray!25} \small{Washington} \\
  \cellcolor{gray!25} (model type) & \cellcolor{gray!25} \footnotesize{CER/WER (\%)} & \cellcolor{gray!25} \footnotesize{CER/WER (\%)} &
    \cellcolor{gray!25} \footnotesize{CER/WER (\%)} \\
  \hline
    \small{Davoudi \& Traviglia \cite{2023DavoudiHDiscrete}} & $6.79/-$ & $4.64/-$ & $\mathbf{3.86}/-$ \\
    \small{(CRNN w. quant. mod.)} & & \\
    \hline
    \small{Abdallah \textit{et al.} \cite{2020AbdallahAAttention}} & $7.25/\mathbf{23.0}$ & $-$ & $8.70/21.50$  \\
    \small{(GCRNN w. attention)} & & \\
    \hline
    \small{Poulos \textit{et al.}} \cite{2021PoulosJCharacter} & $12.7/-$ & $4.7/-$ & $-$ \\
    \small{(CRNN w. attention)} & & \\
    \hline
    \small{Bensouilah \textit{et al.} \cite{2023BensouilahMGmlp}} & $7.6/-$ & $1.58/-$ & $-$ \\
    \small{(gMLP)} & & \\
    \hline
    \small{Shi \textit{et al.} \cite{2016ShiBAn} $\dagger$} & $5.84/37.99^\ast$ & $1.37/6.08$ & $8.25/31.68$ \\
    \small{(CRNN)} & & \\
    \hline
    \small{Shi \textit{et al.} $\dagger$} & $\mathbf{4.09}/30.12$ & $1.26/5.61$ & $7.61/31.08$\\
    \small{w. RL} & & \\
    \hline
    \small{Puigcerver \cite{2017PuigcerverJAre} $\dagger$} & $5.11/33.10$ & $1.67/6.94^\ast$ & $11.48/35.53$\\
    \small{(CRNN)} & & \\
    \hline
    \small{Puigcerver $\dagger$} & $4.53/31.39$ & $1.48/6.45$ & $6.34/23.37$ \\
    \small{w. RL} & & \\ 
    \hline
    \small{Retsinas \textit{et al.} \cite{2022RetsinasGBest} $\dagger$} & $4.68/33.60$ & $1.24/5.33$ & $5.00/20.89$ \\
    \small{(CRNN)} & & \\
    \hline
    \small{Retsinas \textit{et al.} $\dagger$} & $4.62/32.93$ & $\mathbf{1.17/5.21}$ & $4.55/\mathbf{19.52}$ \\
    \small{w. RL} & & \\
    \hline
    \small{Coquenet \textit{et al.} \cite{2022CoquenetDEnd} $\dagger$} & $5.98/38.47$ & $1.35/5.49$ & $5.41/22.35$ \\
    \small{(FCN)} & & \\
    \hline
    \small{Coquenet \textit{et al.} $\dagger$} & $5.90/37.68$ & $1.31/5.88$ & $5.00/20.12$ \\
    \small{w. RL} & & \\
   \hline
\end{tabular}
\caption{Recognition results on the IAM-HisDB datasets. $\dagger$: the model was re-implemented. $-$: data is not available. $^\ast$: a lower learning rate of $1E-04$ was used.
}\label{tab:results_IAM_HisDB}
\end{table*}

For the comparison with the baselines, the architectures were initially trained to achieve the baseline metrics. Training settings across the architectures were kept consistent, with a single exception. We used the Adam optimiser \cite{2014KingmaDPAdam} with an initial learning rate of $1E-3$. In the case of Shi \textit{et al.} \cite{2016ShiBAn}'s architecture, we used a reduced initial learning rate of $1E-4$ since the original value of $1E-3$ did not lead to good performance. Additionally, for all the models, the learning rate was adjusted by a multiplicative factor of $0.1$ after $80$ epochs, when the validation metric was not decreasing further. All models were trained for $400$ epochs in total.

Subsequently, we retrained the same models from scratch, integrating the RL module and evaluating their new performance. We set the loss hyperparameter $\beta$ to $0.1$ as in \cite{2023FerroSExploiting}, while we varied the hyperparameter $\gamma$ over the set $\left\{1E-3, 1E-2, 1E-1\right\}$, using a batch size of $20$. The hyperparameter for the second CTC term in the loss function of Retsinas \textit{et al.} \cite{2022RetsinasGBest} was kept to $0.1$, consistent with its setting in the original paper. Notably, we were able to achieve better results, compared to \cite{2023FerroSExploiting} with a consistently low number of iterations, \textit{i.e.}, $\mathcal{T} = 1, \dots, 5$ (\textit{cf.} next section).
\begin{table*}[tp]
\centering
\begin{tabular}{lcc}
  \hline
  \cellcolor{gray!25} \small{Model authors} (model type) & \cellcolor{gray!25} \small{Simple} & \cellcolor{gray!25} \small{IAM} \\
  \cellcolor{gray!25} & \cellcolor{gray!25} \small{post-proc.} & \cellcolor{gray!25} \footnotesize{CER/WER (\%)}\\
  \hline
    \small{Pham \textit{et al.} \cite{2014PhamVDropout} (MDRNN)} & $\times$ & $10.80/35.10$ \\
    \hline
    \small{Moysset \& Messina \cite{2019MoyssetBAre} (2D-LSTM)} & $\times$ & $8.9/29.3$ \\
    \hline
    \small{Coquenet \textit{et al.} \cite{2020CoquenetDRecurrence} (GFCN)} & $\times$ & $7.99/28.61$ \\
    \hline
    \small{Bluche \cite{2016BlucheTJoint} (MDRNN)} & $\times$ & $7.9/24.6$ \\
    \hline
    \small{Kang \textit{et al.} \cite{2022KangLPay} $^\ast$ (Transformer)} & $\times$  & $7.62/24.54$ \\
    \hline
    \small{Barrere \textit{et al.} \cite{2022BarrereKA} $^\ast$ (Transformer) } & $\times$ & $5.70/18.86$ \\
    \hline
    \small{Cascianelli \textit{et al.} \cite{2022CascianelliSBoosting}} & $\times$ & $7.5/26.9$ \\
    \small{CRNN w. def. conv.} & \\
    \hline
    \small{Cascianelli \textit{et al.} \cite{2022CascianelliSBoosting}}  & $\times$ & $6.8/24.7$ \\
    \small{(CRNN w. def. conv. \& diff. RNN)} & \\
    \hline
    \small{Shi \textit{et al.} $\dagger$} & $\checkmark$ & $7.10/21.42$ \\
    \hline
    \small{Shi \textit{et al.} $\dagger$ w. RL} & $\times$ & $6.68/21.98$ \\
    \hline
    \small{Shi \textit{et al.} $\dagger$ w. RL} & $\checkmark$ & $6.50/20.22$\\
    \hline
    \small{Puigcerver \cite{2017PuigcerverJAre} $\dagger$ (CRNN)} & $\times$ & $12.39/32.20$ \\
    \hline
    \small{Puigcerver $\dagger$} & $\checkmark$ & $12.33/30.18$ \\
    \hline
    \small{Puigcerver $\dagger$ w. RL} & $\times$ & $10.21/27.74$ \\
    \hline
    \small{Puigcerver $\dagger$ w. RL} & $\checkmark$ & $10.20/26.06$ \\
    \hline
    \small{Retsinas \textit{et al.} \cite{2022RetsinasGBest} $\dagger$ (CRNN)} & $\times$ & $6.03/19.49$ \\
    \hline
    \small{Retsinas \textit{et al.} $\dagger$} & $\checkmark$ & $5.99/18.27$\\
    \hline
    \small{Retsinas \textit{et al.} $\dagger$ w. RL} & $\times$ & $\mathbf{5.40/17.85}$ \\
    \hline
    \small{Retsinas \textit{et al.} $\dagger$ w. RL} & $\checkmark$ & $\mathbf{5.33/16.67}$ \\
    \hline
    \small{Coquenet \textit{et al.} \cite{2022CoquenetDEnd}$\dagger$ $^{\ast\ast}$ (FCN)} & $\times$ & $6.47/21.97$ \\
    \hline
    \small{Coquenet \textit{et al.} $\dagger$} & $\checkmark$ & $6.27/19.94$ \\
    \hline
    \small{Coquenet \textit{et al.} $\dagger$ w. RL} & $\times$ & $6.13/20.82$ \\
    \hline
    \small{Coquenet \textit{et al.} $\dagger$ w. RL} & $\checkmark$ & $5.89/18.85$ \\
    \hline
\end{tabular}
\caption{Recognition results on the IAM dataset. $\dagger$: the model was reimplemented. $^\ast$: only the results derived from the training data, without incorporating any additional synthetic data, were considered to ensure a fair comparison.
$^{\ast\ast}$: the model has better performance in the original paper, due to a more extensive image augmentation (we use the same for all models).
}\label{tab:results_IAM}
\end{table*}

\section{Experimental Results}
\label{sec:exp_results}
To guarantee a fair comparison, we considered only models that do not incorporate any Language Model (LM) in their transcription process. The main objective was to improve the recognition capabilities of the models, particularly during the initial recognition phase, before applying any post-processing. The performance of the HTR models was evaluated using the standard metrics of Character Error Rate (CER) and Word Error Rate (WER).

The CER is given by the following formula
\begin{equation}\label{eq:cer}
    CER = \frac{\sum_{i=1}^n d(\hat{y}_i, y_i)}{\sum_{i=1}^n |y_i|},
\end{equation}
where $d(\cdot)$ is the Levenshtein distance \cite{1966LevenshteinVIBinary} calculated between the predicted character sequence $\hat{y}$ and the ground truth $y$, $|\cdot|$ is the number of characters in the sequence, and $n$ is the number of sample sequences. In essence, the CER represents the fraction of the number of substituted, deleted and inserted elements in the sequence with respect to the number of elements in the reference/target sequence. WER is computed with the same formula, at the word level.

\subsection{Quantitative Analysis}
\subsubsection{Model Comparison}
Tab.~\ref{tab:results_IAM_HisDB} reports the results on the historical datasets. As can be seen, the application of RL (``w. RL'' models in the table) consistently lowers the CER and WER metrics of the corresponding baseline models. Additionally, some RL-trained models set new state-of-the-art results in at least one of the two considered metrics, mostly in both. In the case of the Saint Gall and Washington datasets, the CER and WER results for the best model do not coincide. We attribute this variability to the small size of the two training sets (respectively, $468$ and $325$ samples), for which many models present already high performance.
\begin{figure*}[!t]
    \centering
    \includegraphics[width=0.40\textwidth]{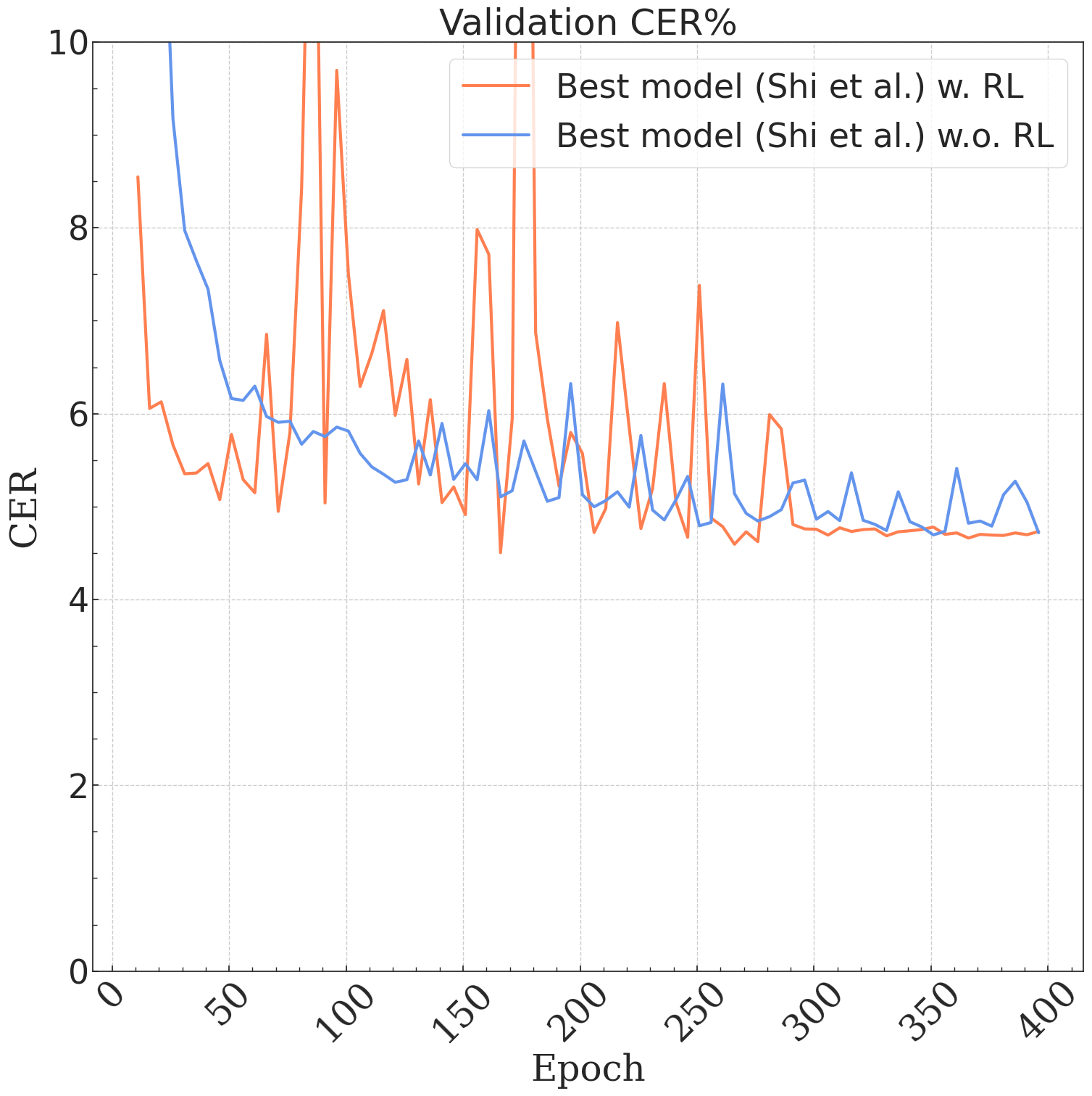}
    \includegraphics[width=0.40\textwidth]{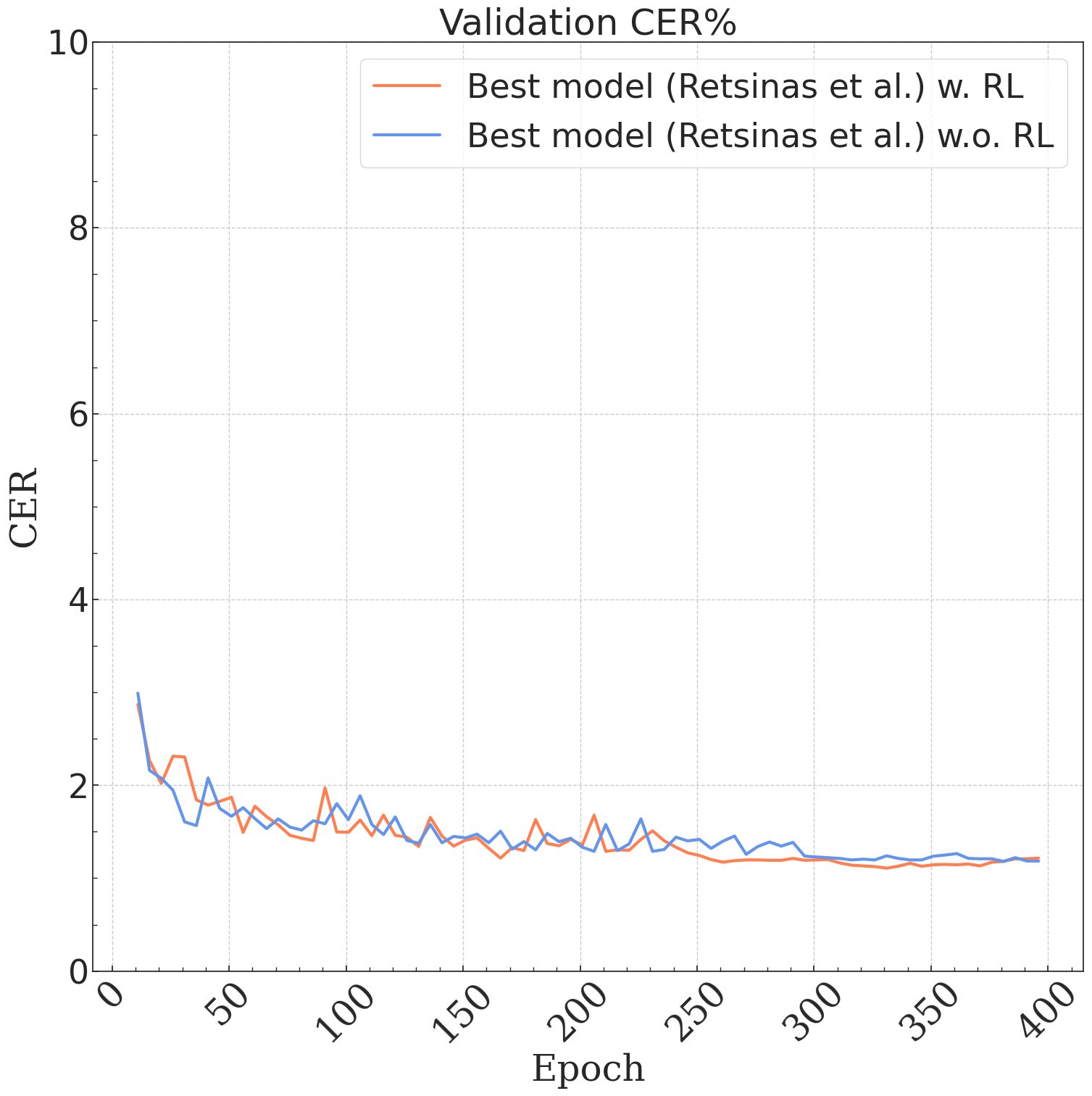}
    \includegraphics[width=0.40\textwidth]{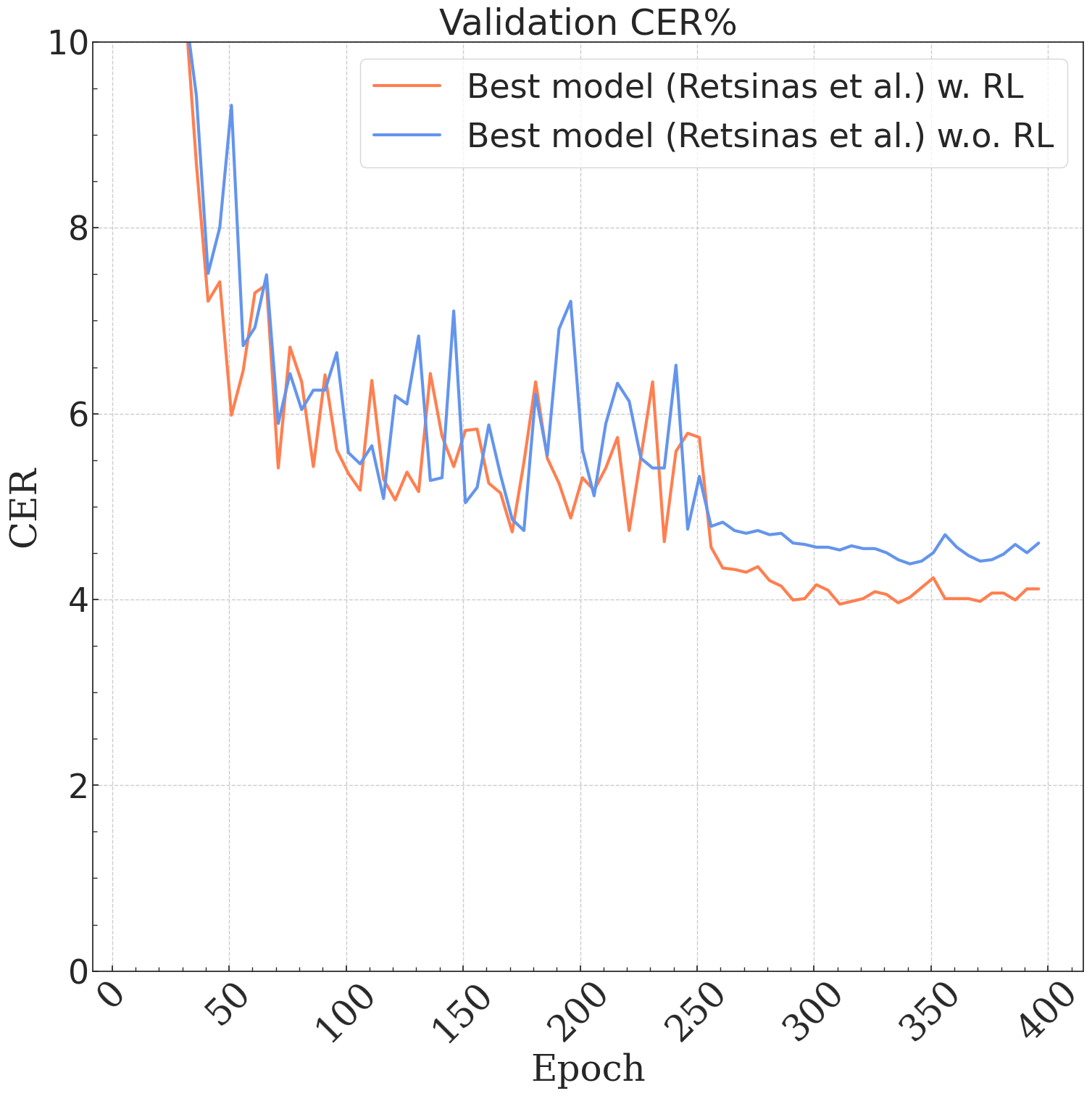}
    \includegraphics[width=0.40\textwidth]{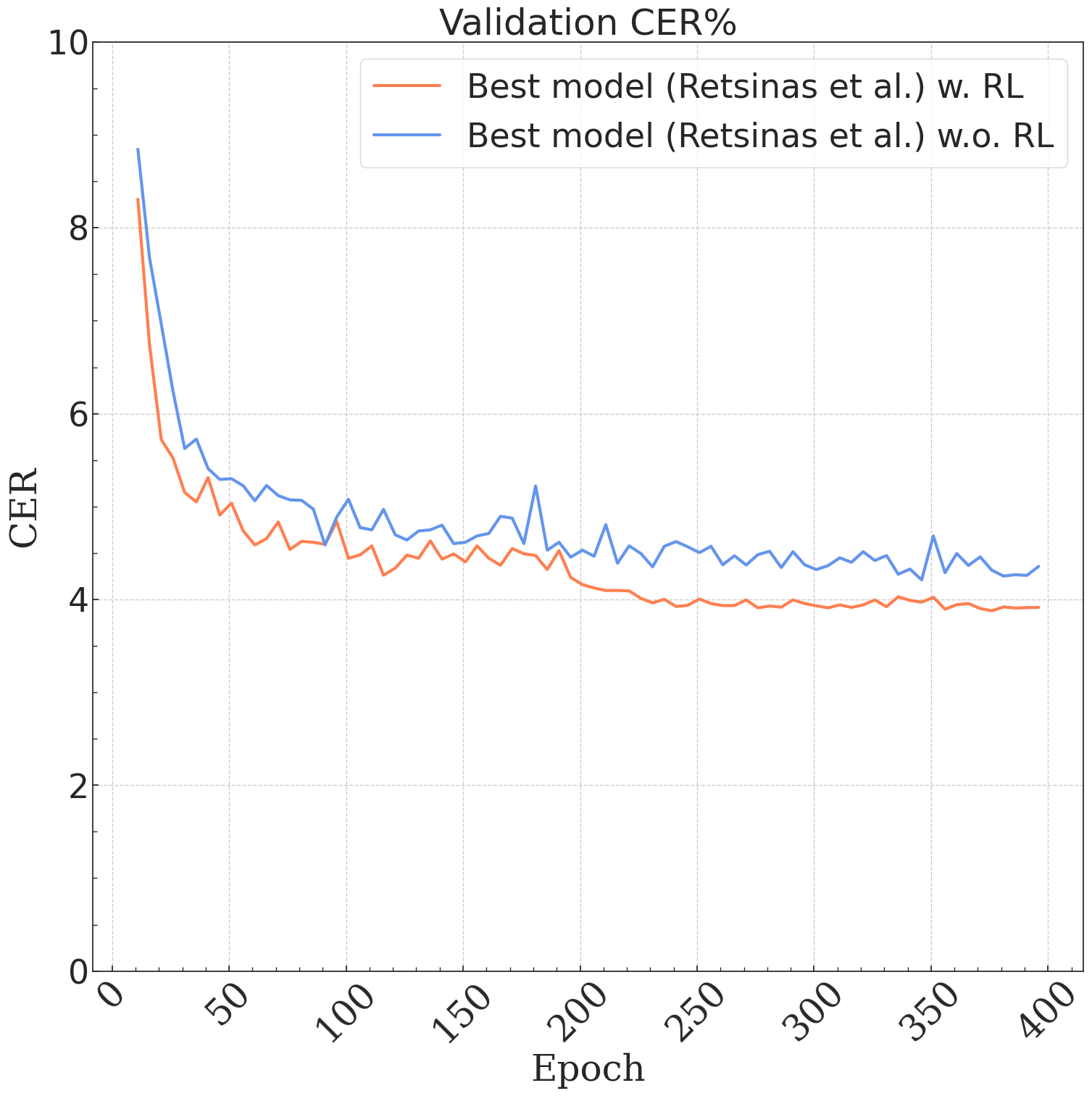}
    \caption{Validation CER curves of the best model with and without the application of RL for Saint Gall (top-left), Parzival (top-right), Washington (bottom-left), IAM (bottom-right).}
    \label{fig:top_val_cer}
\end{figure*}

Tab.~\ref{tab:results_IAM} details the findings on the IAM dataset, with similar outcomes. Incorporating the RL module allows all models to achieve enhanced performance. 

\subsubsection{Simple Post-processing}\label{ssec:simple_post_proc}
To further enhance the performance of our models we applied a straightforward but effective post-processing on the IAM dataset. The post-processing is based on the Levenshtein distance at character level. Considering that the model usually produces outcomes closely matching the correct words, we replaced the predicted words with their nearest neighbours searched in a vocabulary. In particular, if the Levenshtein distance between the predicted word and its nearest neighbour was below a specified threshold, we replaced it. We composed our vocabulary by integrating the training set and an external source (again, the corpora in \citep{2009BirdSNatural}).

Tab.~\ref{tab:results_IAM} shows the results of the post-processing applied to the models trained. Taking into account the transcriptions provided by the optimal model (Retsinas \textit{et al.}) the CER is effectively lowered from $5.40\%$ to $5.33\%$, reaching an improved performance. Such an effect is consistently observed across all other tested models as well.
\begin{table}[!t]
    \centering
       \begin{tabular}{lcccccccc}
       \toprule
        \textbf{Model} & \multicolumn{2}{c}{\textbf{OOV (\%)}} \\
         & \textbf{w.o. RL} & \textbf{w. RL} \\
        \midrule
        Shi \textit{et al.} & $11.09$ &  $\mathbf{12.00}$ \\
        Puigcerver & $12.11$ & $\mathbf{12.44}$ \\
        Retsinas \textit{et al.} & $10.82$ & $\mathbf{11.02}$ \\
        Coquenet \textit{et al.} & $\mathbf{10.57}$ & $10.23$ \\     
        \bottomrule
    \end{tabular} 
    \caption{Out-of-vocabulary performance on baseline models (w.o. RL) and the models with relaxation labelling (w. RL) on the IAM dataset. OOV denotes the percentage of words present in an external vocabulary and not in IAM. In this context, a word is defined as any sequence of characters delimited by spaces, removing punctuation.}\label{tab:statistics_baseline_and_relab}
\end{table}

\subsubsection{Learning Curves}
\begin{table*}[t!]
    \centering
    \begin{tabular}{lcccccc}
        \toprule
        \textbf{Dataset} & $\mathbf{\mathcal{T}}$ & $\mathbf{\gamma}$ & \textbf{Val.} &  \textbf{Test} &  \textbf{Val.} &  \textbf{Test} \\
         & & & \textbf{CER} &  \textbf{CER} &  \textbf{WER} &  \textbf{CER} \\
        \midrule
        \multicolumn{7}{c}{Shi \textit{et al.}} \\
        \midrule
        IAM & $3$ & $1E-1$ & $4.59$ & $6.68$ & $16.21$ & $21.98$ \\
        Parzival & $5$ & $1E-3$ & $1.13$ & $1.26$ & $4.78$ & $5.61$ \\
        Saint Gall & $3$ & $1E-1$ & $3.95$ & $4.09$ & $29.13$ & $30.12$ \\
        Washington & $5$ & $1E-1$ & $5.40$ & $7.61$ & $23.60$ & $31.08$ \\
        \midrule
        \multicolumn{7}{c}{Puigcerver} \\
        \midrule
        IAM & $2$ & $1E-3$ & $7.95$ & $10.20$ & $22.76$ & $27.74$ \\
        Parzival & $4$ & $1E-3$ & $1.49$ & $1.48$ & $6.58$ & $6.45$ \\
        Saint Gall & $3$ & $1E-3$ & $4.43$ & $4.53$ & $30.32$ & $31.39$ \\
        Washington & $3$ & $1E-1$ & $6.13$ & $6.34$ & $24.17$ & $23.37$ \\
        \midrule
        \multicolumn{7}{c}{Retsinas \textit{et al.}} \\
        \midrule
        IAM & $2$ & $1E-1$ & $3.73$ & $5.40$ & $13.40$ & $17.85$ \\
        Parzival & $2$ & $1E-3$ & $1.10$ & $1.17$ & $4.59$ & $5.21$ \\
        Saint Gall & $2$ & $1E-1$ & $4.41$ & $4.62$ & $31.62$ & $32.93$ \\
        Washington & $2$ & $1E-1$ & $3.92$ & $4.55$ & $18.23$ & $19.52$ \\
        \midrule
        \multicolumn{7}{c}{Coquenet \textit{et al.}} \\
        \midrule
        IAM & $4$ & $1E-2$ & $4.41$ & $6.13$ & $16.16$ & $20.82$ \\
        Parzival & $1$ & $1E-3$ & $1.19$ & $1.31$ & $4.91$ & $5.88$ \\
        Saint Gall & $2$ & $1E-3$ & $4.50$ & $5.90$ & $30.62$ & $37.68$ \\
        Washington & $2$ & $1E-3$ & $4.07$ & $5.00$ & $19.04$ & $20.12$ \\
        \bottomrule
    \end{tabular}
    \caption{Best hyperparameter configuration for $\mathcal{T}$ and $\gamma$, together with the obtained validation and test CER and WER for the four tested models.}\label{tab:best_hyperp_shi}
\end{table*}

In Fig.~\ref{fig:top_val_cer}, we report the validation CER curves of the best-performing model over the epochs, in the four tested datasets. Since the curves of the Saint Gall and the Parzival, were not showing a clear difference between the performance with and without RL, we ran the training for an additional $400$ epochs. In all four cases, the RL processes drive the network towards consistent labellings effectively lowering the validation CER and WER, with respect to the corresponding baselines. Again, Saint Gall and Washington show a high variability and the convergence to a minimum is more noisy, due to the limited size of these datasets.

\subsubsection{Out-Of-Vocabulary Words}
For the IAM dataset, the only one in a modern language, we carried out an analysis of the effects of the RL module in generating correct out-of-vocabulary words that were not present in the original dataset and would be therefore marked as errors. This evaluation sought to assess the capacity of RL to not only correct existing words but also to achieve a more precise alignment with a wider general vocabulary.

Tab.~\ref{tab:statistics_baseline_and_relab} presents the percentage of distinct words in an external vocabulary composed by the corpora in \citep{2009BirdSNatural} where the words in the IAM vocabulary (considering all the splits) were removed. These statistics are provided for both the baseline and the model enhanced with RL. For almost all the cases, when using RL, we have an increase in such percentage, meaning that the produced words are coherent with the English language. In the case of Coquenet \textit{et al.}, even though such percentage decreases, RL is anyway improving the overall CER, as shown previously in Tab~\ref{tab:results_IAM}. 

\subsubsection{Hyperparameters Configurations}
Tab.~\ref{tab:best_hyperp_shi} show the best hyperparameters configurations for $\mathcal{T}$ and $\gamma$. Regarding $\mathcal{T}$, in general, a number of iterations lower than $5$ gives the best results, thus keeping the computational training cost of the RL module low. This value is more than three times lower than the number of iterations in \cite{2023FerroSExploiting}. This phenomenon is due to the effect of the $\ell_1$ sparsification term that cancels out the effect of small, noisy compatibilities, therefore accelerating the convergence of RL. In the case of $\gamma$, the optimal selection varies with the architecture and the dataset employed.

\subsection{Qualitative Analysis}
\begin{table*}[!t]
\centering
\caption{Examples of refinement using the RL module for all the datasets. Grey-filled bounding boxes highlight errors.
The sample from the IAM dataset reports a case of character substitution, the one from the Parzival dataset presents a case of insertion, while those from the Saint Gall and Washington datasets highlight examples of deletion.}\label{tab:qual_an} \centering
\begin{tabular}{cc}
\Xhline{2pt}
  \cellcolor{gray!25} Sample &  \includegraphics[width=0.5\textwidth]{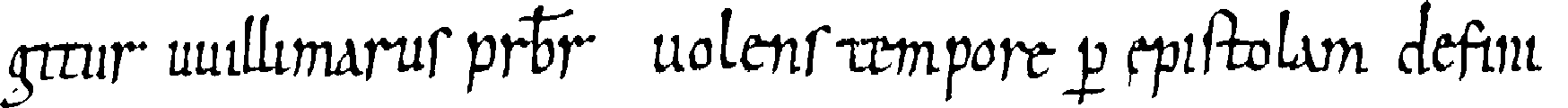} \\ 
  \cellcolor{gray!25} (Saint Gall) & \\
  \hline
  \cellcolor{gray!25} GT & gitur willimarus prbr volens tempore p epistolam defini \\
  \hline
  \cellcolor{gray!25} w.o. RL &
  \begin{tikzpicture}
  \path
  ( -7.65,0.97) node [shape=rectangle, fill=gray!35, draw, opacity=0.65] {vv}
  ( -8.5,0.98) node [] {gitur w}
  ( 0.75,0.98) node [shape=rectangle, fill=gray!35, draw, opacity=0.65] { }
  ( -3.4,0.98) node [] {illimarus prbr volens tempore p epistolam de}
  ( 1.15,1) node [] {fini};
  \end{tikzpicture}\\
  \hline
  \cellcolor{gray!25} w. RL & gitur willimarus prbr volens tempore p epistolam defini \\
\Xhline{2pt} & \\[-1.5ex]
\Xhline{2pt}
  \cellcolor{gray!25} Sample &  \includegraphics[width=0.20\textwidth]{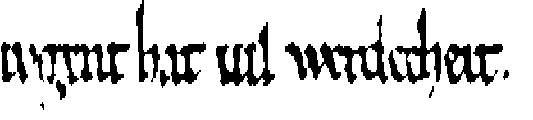} \\
  \cellcolor{gray!25} (Parzival) & \\
  \hline
  \cellcolor{gray!25} GT & ivgent hat uil werdecheit. \\
  \hline
  \cellcolor{gray!25} w.o. RL &
 \begin{tikzpicture}
 \path
    ( -3.68,0.98) node [shape=rectangle, fill=gray!35, draw, opacity=0.65] {}
    ( -3.75,1) node [] {i}
    ( -1.52,0.97) node [] {gent hat uil werdecheit.};
 \end{tikzpicture}\\
\hline
  \cellcolor{gray!25} w. RL & ivgent hat uil werdecheit. \\
\Xhline{2pt} & \\[-1.5ex]
\Xhline{2pt} 
  \cellcolor{gray!25} Sample &  \includegraphics[width=0.40\textwidth]{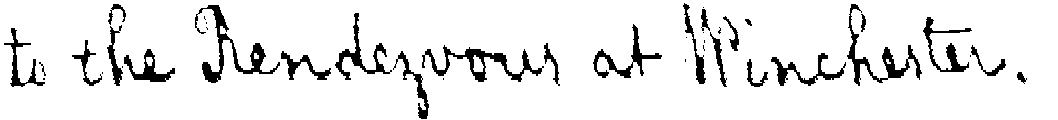} \\
   \cellcolor{gray!25} (Washington) & \\
  \hline
  \cellcolor{gray!25} GT & to the Rendezvous at Winchester. \\
  \hline
  \cellcolor{gray!25} w.o. RL &
 \begin{tikzpicture}
 \path
  ( -1.28,0.965) node [shape=rectangle, fill=gray!35, draw, opacity=0.65] {s}
  ( -3.5,1) node [] {to the Rendezvous at W}
  ( -0.34,0.99) node [] {inchester.};
\end{tikzpicture}\\
\hline
  \cellcolor{gray!25} w. RL & to the Rendezvous at Winchester. \\
\Xhline{2pt} & \\[-1.5ex]
\Xhline{2pt} 
  \cellcolor{gray!25} Sample &  \includegraphics[width=0.40\textwidth]{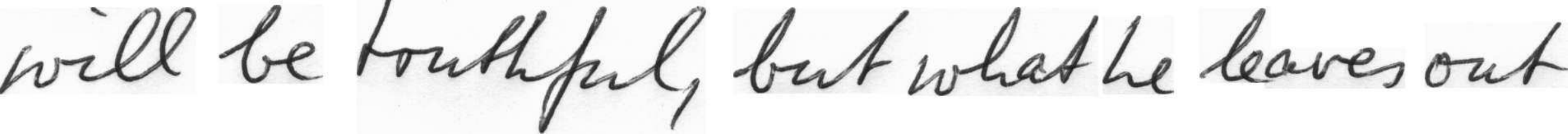} \\
  \cellcolor{gray!25} IAM & \\
  \hline
  \cellcolor{gray!25} GT & will be truthful , but what he leaves out \\
  \hline
  \cellcolor{gray!25} w.o. RL &
 \begin{tikzpicture}
 \path
  ( -3.08,0.96) node [shape=rectangle, fill=gray!35, draw] {o}
  ( -3.9,1) node [] {will be t}
  ( -0.18,0.98) node [] {uthful , but what he leaves out};
\end{tikzpicture}\\
\hline
  \cellcolor{gray!25} w. RL &
will be truthful , but what he leaves out \\
\Xhline{2pt}
\end{tabular}
\end{table*}

Tab.~\ref{tab:qual_an} reports one case from each dataset where the RL module increases transcription accuracy. It can be noted that the model is capable of performing all types of modifications to the text line.

\section{Conclusions and Future Work}\label{sec:conc}
In this paper, we have demonstrated that learnable relaxation labelling processes greatly enhance the generalization capabilities of well-established baseline architectures for HTR. We have also shown that RL benefits from a sparsification procedure applied to the compatibility matrix, which accelerates the process’s convergence to a consistent labelling.
In some cases, the RL-enhanced models compete with or even beat recent transformer-based architectures, despite being substantially smaller in size. RL plays a crucial role in driving the network towards consistent labellings, improving the overall performance of the system in terms of both CER and WER. Additionally, in the specific case of modern English handwriting recognition (the sole scenario where an external vocabulary was available) we were able to assess that RL also contributes to increasing the number of out-of-vocabulary words, thus indicating enhanced linguistic coherence. Finally, we have shown that a straightforward post-processing step can further enhance the overall performance of the trained models.

As previously mentioned, we think there is an affinity between the RL processes and the self-attention module of the transformer architecture. In light of this, we aim to conduct a detailed comparison between the two methods. Additionally, this work covers only contextual information at the text-line level. In the future, we aim to consider broader contexts, such as sentence- or paragraph-level, to further enhance the recognition accuracy.

\section{Acknowledgements}
The authors would like to express their gratitude to Dr. Marco Fiorucci and Dr. Marina Ljubenovic for their valuable insights and thorough review of this paper. Their constructive feedback and suggestions have significantly contributed to the improvement and clarity of the manuscript.

\bibliographystyle{elsarticle-num} 
\bibliography{refs}





\end{document}